\documentclass{ieeeaccess}
\usepackage{cite}
\usepackage{amsmath,amssymb,amsfonts}
\usepackage{makeidx}
\usepackage{balance}
\usepackage{tipa}
\usepackage{subfig}
\usepackage{float}
\usepackage{tabularx}
\usepackage{makecell}
\usepackage{multirow}
\usepackage[ruled,lined]{algorithm2e}

\usepackage{algpseudocode}
\usepackage{hyperref}
\usepackage{caption}
\usepackage{graphicx}
\usepackage{textcomp}
\usepackage{placeins}
\usepackage{color}
\usepackage{xcolor}
\usepackage{soul}
\usepackage{letltxmacro}
\usepackage{rotating}
\usepackage{wrapfig}
\usepackage{nth}
\usepackage{tablefootnote}
\DeclareMathOperator{\atantwo}{atan2}

\DeclareMathOperator{\cre}{Re}
\DeclareMathOperator{\cim}{Im}

\def\BibTeX{{\rm B\kern-.05em{\sc i\kern-.025em b}\kern-.08em
    T\kern-.1667em\lower.7ex\hbox{E}\kern-.125emX}}
\begin{document}
\history{Date of publication xxxx 00, 0000, date of current version xxxx 00, 0000.}
\doi{10.1109/ACCESS.2017.DOI}

\title{LSDNet: Trainable Modification of LSD Algorithm for Real-Time Line Segment Detection}
\author{\uppercase{Lev Teplyakov}\authorrefmark{1},
\uppercase{Leonid Erlygin\authorrefmark{1,2},
Evgeny Shvets}\authorrefmark{1}}
\address[1]{Institute for Information Transmission Problems, Russian Academy of Sciences, 119991 Moscow, Russia}
\address[2]{Moscow Institute of Physics and Technology, 117303 Moscow, Russia}
\tfootnote{This work was supported by the Russian Science Foundation (Project No. 20-61-47089)}

\markboth
{Author \headeretal: Preparation of Papers for IEEE TRANSACTIONS and JOURNALS}
{Author \headeretal: Preparation of Papers for IEEE TRANSACTIONS and JOURNALS}

\corresp{Corresponding author: Lev Teplyakov (e-mail: teplyakov@iitp.ru).}

\begin{abstract}
As of today, the best accuracy in line segment detection (LSD) is achieved by algorithms based on convolutional neural networks -- CNNs.
Unfortunately, these methods utilize deep, heavy networks and are slower than traditional model-based detectors.
In this paper we build an accurate yet fast CNN-based detector, LSDNet, by incorporating a lightweight CNN into a classical LSD detector.
Specifically, we replace the first step of the original LSD algorithm -- construction of line segments heatmap and tangent field from raw image gradients -- with a lightweight CNN, which is able to calculate more complex and rich features.
The second part of the LSD algorithm is used with only minor modifications.
Compared with several modern line segment detectors on standard Wireframe dataset, the proposed LSDNet provides the highest speed (among CNN-based detectors) of $214$ FPS with a competitive accuracy of $78$ $F^H$.
Although the best-reported accuracy is $83$ $F^H$ at $33$ FPS, we speculate that the observed accuracy gap is caused by errors in annotations and the actual gap is significantly lower.
We point out systematic inconsistencies in the annotations of popular line detection benchmarks -- Wireframe  and  York Urban, carefully reannotate a subset of images and show that (i) existing detectors have improved quality on updated annotations without retraining, suggesting that new annotations correlate better with the notion of correct line segment detection; (ii) the gap between accuracies of our detector and others diminishes to negligible $0.2$ $F^H$, with our method being the fastest.

\end{abstract}

\begin{keywords}
Convolutional neural networks, edge detection, line segment detection, U-net, LSD
\end{keywords}

\titlepgskip=-15pt

\maketitle

\begin{figure*}[htb]
\centering\includegraphics[width=0.9\textwidth]{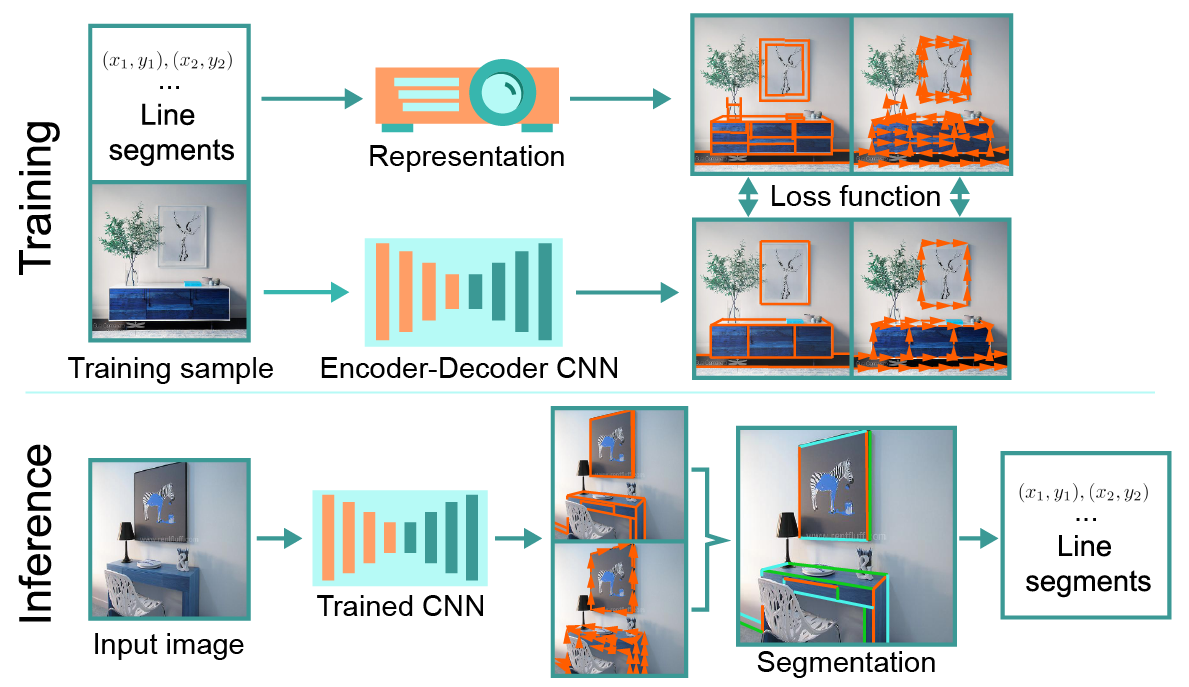}
\caption{Overview of the proposed approach. A lightweight neural network predicts line segment mask and tangent vector field, which are then clustered and each cluster is represented as a line segment.}
\label{fig:approach}
\end{figure*}

\section{Introduction}
\PARstart{A}{utomatic} general-purpose line segment detection is a long-standing computer vision problem of high practical importance.
Line segment detectors are exploited to construct an intermediate representation of image contents in visual recognition systems over a wide range of applications, such as autonomous vehicle localization \cite{andrade2018novel, hoang2016road, shipitko2020linear}, infrastructure maintenance with an UAV \cite{ceron2014power, cao2012automatic}, document recognition \cite{shemiakina2018method, zhukovsky2017segments}.

Traditionally, the problem of line segment detection was approached with so-called model-based algorithms \cite{von2008lsd, akinlar2011edlines, guerreiro2012connectivity, xu2014accurate}.
These algorithms operate by searching an image for elements that satisfy an explicit definition of a salient line segment, for example, ``line segment is a strip-like set of image pixels with similar gradients'' \cite{von2008lsd, akinlar2011edlines} or ``an image region is a line segment if its contour map triggers a peak in Hough space'' \cite{guerreiro2012connectivity, xu2014accurate}.
These algorithms typically have the benefits of being fast and having interpretable parameters. However, they may miss the segments that are salient for human, but for some reason don't match the exact explicit implemented definition. They are also prone to over-segmentation (splitting a single segment into parts) and sometimes demand nontrivial problem-specific postprocessing \cite{hamid2016lsm, yu2020plsd}.

The troubling problem of formulating an explicit criterion that matches the human expectation of what exactly constitutes a ``salient line segment'' can be avoided by manually annotating images and training a CNN, which then learns an implicit algorithm from data samples. This is an approach that yields the best accuracy in line segment detection task today \cite{xue2019learning, xue2020holistically, lin2020deep, huang2020tp, zhou2019end, huang2019wireframe, gu2021towards}. Skipping ahead, let us note that such annotation is not a simple task either - existing datasets on line segment detection have numerous and sometimes extreme internal inconsistencies - probably caused by the inherent ambiguity of the task, lack of clear labeling instructions and the tediousness of the task, leading to missed segments.

From a technical perspective, there also is a challenge in designing a CNN-based line detector. 
The typical solution is that CNN constructs an intermediate representation - encoding - which is then converted into a set of answers by some hand-crafted algorithm.
Object detection networks have solved this problem by using so-called anchors \cite{bochkovskiy2020yolov4} and, later, more simple anchorless detectors such as FCOS \cite{tian2019fcos}.
But line segment detectors need alternative encodings, suitable not for bounding boxes, but for line segments.
The encodings should effectively deal with the fact that the line segments to be detected in a typical image intersect with each other a lot, while 
encodings for bounding boxes are effective only when ``overlapping mostly happens between objects with considerably different sizes'' \cite{tian2019fcos}, making them hard to exploit for line segment detection.
While object detectors encondings - after many iterations of refinement - have become fast and elegant, we believe that intermediate representations of most line detector used today are still either imprecise, slow or unintuitive.

So whether it is the complexity of the interpreter or the sheer weight of the CNN backbone, CNN-based detectors that outperform the traditional ones in accuracy are also computationally harder \cite{gu2021towards}. Their complexity limits the scope of application of such algorithms in cases where speed, energy consumption, or hardware price are critical. 

In this work we propose a fast yet accurate CNN-based line segment detection algorithm, LSDNet, built on the basis of a widely used model-based detector, LSD \cite{von2008lsd}.
LSDNet overview is presented in Figure \ref{fig:approach}.
The first step of LSD is the calculation of image gradient's orientation and magnitude.
We view this step as an estimation of an intermediate representation,
composed of line segments' heatmap, estimated as gradient's magnitude, and tangent field, estimated as gradient's orientation.
We substitute this step with a lightweight CNN to generate the heatmap and tangent field from a more diverse and complex set of features than a simple gradient; the second step, conversion the intermediate representation into a set of line segments, is taken from the LSD almost as-is.
This substitution boosts the LSD accuracy and simplifies the postprocessing due to more accurate heatmap and tangent field. The heatmap and tangent field generation is a relatively simple task - the answer can be correctly inferred from the local context - which allows to use a lightweight CNN.

We compare LSDNet and a selection of existing detectors: LSD \cite{von2008lsd}, L-CNN\cite{zhou2019end}, HAWP\cite{xue2020holistically}, TP-LSD\cite{huang2020tp}, M-LSD and M-LSD-tiny\cite{gu2021towards} and show that LSDNet provides competitive accuracy of $78$ $F^H$ score on Wireframe dataset while being the fastest:  $214$ FPS for $288\times288$ input on conventional hardware. LSDNet ourperforms the second fastest approach, M-LSD-tiny\cite{gu2021towards}, both in accuracy and speed.

We discuss considerable inconsistencies (section \ref{sec:dataset}) in ground truth labeling of current benchmarks for line segment detection accuracy - datasets Wireframe \cite{huang2018learning} and YorkUrban \cite{denis2008efficient}. The labeling in these datasets is inconsistent not only between images: within the very same image many segments, similar in appearance, are often marked up differently - some as a positives, others as negatives. We speculate that these datasets in their current state are flawed for assessing the accuracy of general purpose line segment detectors.

So, in addition to measuring the accuracy of our detector on these datasets, we select and reannotate a part of Wireframe dataset - consisting mainly of simple images with less ambiguity - and show that the reannotated subset correlates better with conventional notion of correct line segment detection. 
Specifically, it narrows the accuracy gap between LSD and CNN-based approaches, and makes the gap between the proposed LSDNet and L-CNN \cite{zhou2019end}, one the most accurate CNN-based approaches, negligible.

\section{Related Work}
In this section we cover the model-based LSD algorithm \cite{von2008lsd} and the existing CNN-based approaches to the problem of line segment detection.

\subsection{LSD Detector}

LSD is one of the most popular general-purpose line segment detectors and serves as a common baseline for CNN-based algorithms both in accuracy and speed \cite{xue2020holistically, lin2020deep, huang2020tp}.

The first step of LSD detector is gradient calculation; then the algorithm builds  so-called line support regions (LSRs) \cite{burns1986extracting}. LSRs are image segments (not to be confused with line segments), spanning actual line segments in an image. They are built by iteratively grouping neighbouring pixels with high gradients' magnitudes and similar orientations. After the formation of initial LSRs, they undergo several steps of filtering and refinement. These include, among others, splitting LSRs of ``hockey stick'' shape - effectively decoupling distinct but merged regions; removal of LSRs with a high deviation of gradients' orientations - possibly false-positives.
Typical resulting LSRs are long, straight and a few pixels thick. Finally, each such  region is individually encoded as a pair of points - effectively, a line segment.

LSD detector is fast, reaching 185 FPS for $320\times320$ images on a conventional CPU, and provides an accuracy of $63.3$ $F^H$ on the Wireframe dataset.

\subsection{CNN-based approaches}

The CNN-based approaches are typically composed of two modules: the CNN itself predicts an intermediate representation, then a postprocessing module reconstructs line segments from this representation.

We consider the design of the intermediate representation to be the key growth point of CNN-based line segment detectors since the desired detector's output - a unknown-size set of line segments with potentially high overlap - is hard to represent as a ``CNN-friendly'' fixed-shape tensor \cite{teplyakov2020line}.
The survey below covers most popular intermediate representations used in existing CNN-based approaches.

The first CNN-based detectors represented line segments in an image as a set of endpoints and their connectivity graph \cite{huang2018learning, zhou2019end}.
The endpoints were detected as local maxima of a CNN-produced heatmap.
The connectivity of the endpoints was deduced either with the help of edge map heuristics \cite{huang2018learning} or with a trainable classifier \cite{zhou2019end}.
To generate the classifier's input, a fixed number of uniform spaced points was sampled from the feature map between the endpoints.
This operation was called LoI pooling \cite{zhou2019end}.

In \cite{xue2019learning} a representation called attraction field was proposed (distance field \cite{huang2019wireframe} is a similar concept).
Line segments were represented as a 2D vector field of translations to a nearest line segment.
The postprocessing step for such a representation required nontrivial line segments extraction from heatmap-like prediction.
Interestingly, the postprocessing of even a perfect attraction field generated from dataset annotations did not provide absolute detection accuracy \cite{xue2019learning} - in other words, such representation is inherently ambiguous.

The covered representations of endpoints and attraction field were combined and modified in \cite{xue2020holistically}.
The proposed CNN predicted both the endpoints and the attraction field, which was enriched to encode the translations to the both segment's endpoints as 4D vector field.
Then the endpoints and the attraction field were used to refine each other.
The refined line segments proposals underwent final verification with the help of LoI pooling and a trainable classifier.
The detector provides state-of-the-art quality of $83$ $F^H$ on Wireframe \cite{huang2018learning} dataset to date, but with low speed of $33$ FPS on GPU.

The recently proposed ``tri-points'' representation \cite{huang2020tp} is focused on speeding up the detector.
A line segment was represented by its center point and two vectors to its endpoints.
It allowed to significantly boost the speed up to $50$ FPS, driven by a much faster postprocessing requiring trivial conversion from ``tri-point'' to a line segment and non-maximum supression.
The CNN itself remained comparably slow.
In \cite{gu2021towards} some further enhancements were proposed.
A lightweight CNN was designed and the training procedure was improved by augmentations and more sophisticated loss function.
It resulted in the fastest CNN-based detector to date with $200$ FPS overall and $241$ FPS for standalone CNN.

\section{Proposed Approach}

A perfect line segments representation should make it possible to design a CNN and a postprocessing module both being fast and accurate.
We argue that in the search for such a representation there is no need to develop a brand new one from scratch; instead, the representation used implicitly by LSD detector - line segments heatmap and tangent field - already possesses all the desired properties.
Indeed, the heatmap and tangent field could be inferred from local image context and does not require the reasoning of complex abstract features, which allows to use a lightweight CNN.
On the other hand, as proven by LSD, the representation could be efficiently postprocessed to actual line segments.
In the next sections we cover each step of the algorithm in detail.

\subsection{CNN}

\textbf{Line Segment Representation}.
We represent a set of line segments in an image as a 2-channel feature map of the same height and width as the image. The first channel denoted by $M$ contains a line segment mask. The second channel denoted as $F$ contains tangent vector field of line segments. Since the values of $F$ are unit vectors, we encode $F$ as one channel feature map.

Let $l = (x_1,y_1, x_2,y_2)$ be a line segment, $\varphi_l$ - segment's level line angle in range $[0,\pi)$, $p = (x,y)$ - an image pixel, $L_p = l_1, l_2, ...$ - set of line segments crossing $p$.
Then if $L_p=\emptyset$, then $M(p)=0$ and $F(p)$ is arbitrary;
otherwise $M(p)=1$, $F(p)=\varphi_{l_1}$.

Note that in case of overlapping segments ($||L_p|| > 1$) $F(p)$ is defined by one arbitrary segment of overlap. This ambiguity probably can be ignored since only about $1\%$ of line segments' pixels lie on overlaps - as measured on Wireframe dataset \cite{huang2018learning}.

\textbf{Loss Function}.
The loss function $L = L_{mask} + \alpha L_{field}$ used to fit the network is composed of two independent weighted terms, one responsible for mask $M$, other - for the tangent vector field $F$.

While the prediction of mask $M$ is a straightforward segmentation problem with $L_{mask}$ being a conventional cross enthropy loss, to correctly estimate error in prediction of the tangent field $F$ we should account for the following property of line segments' level line angles - the distance between angles $0^0$, $10^0$ and $0^0$, $170^0$ should be equal.

\begin{figure}[htbp]
\centering
\begin{minipage}[b]{.49\linewidth}
  \centering
  \centerline{\includegraphics[height=3.0cm]{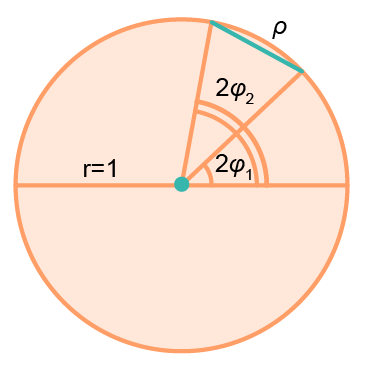}}
  \centerline{(a)}
\end{minipage}
\hfill
\begin{minipage}[b]{.49\linewidth}
  \centering
  \centerline{\includegraphics[height=3.0cm]{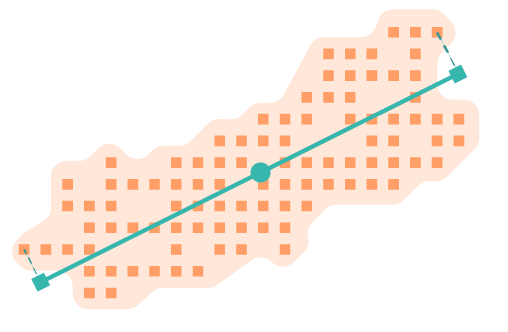}}
  \centerline{(b)}
\end{minipage}
\hfill
\caption{(a) A visualization of the distance function between angles $\varphi_1$, $\varphi_2$. It accounts for the periodicity of angles. (b) Line segment extraction from a region.}
\label{fig:dist}
\end{figure}

This problem can be approached by computing several distances between the original angles and the angles, shifted by $\pm\pi$, and picking the minimum distance \cite{li2021lidar}.
The calculation can be done simpler: let $\varphi_1$, $\varphi_2$ - angles between which the distance is to be computed,  $z_1 = e^{i\varphi_1}, z_2 = e^{i\varphi_2} \in \mathbf{C}$ - the representation of the angles as complex numbers with unit length and phases $\varphi_1$, $\varphi_2$, then
\begin{equation}\label{eq:angledist}
    \rho(\varphi_1,\varphi_2) = ||z_1^2-z_2^2||_2
\end{equation}
This distance function has a geometrical interpretation, illustrated in Fig. \ref{fig:dist}a.
It equals to $\ell^2$-norm of vector difference between unit vectors with phases $2\varphi_1$, $2\varphi_2$.
Turning to the aforementioned example, the doubled phase makes vectors, corresponding to angles $10^0$ and $170^0$, equally close to the horizontal, corresponding to angle $0^0$.

Given the angle distance function $\rho$, the predicted field $F_{p}$, the reference mask $M_t$ and reference field $F_t$,  the tangent vector field loss $L_{field}$ is defined as
\begin{equation}
    L_{field} = \frac{1}{\sum_p M_{t}(p)}\sum_{p: M_{t}(p) = 1}\rho^2(F_{t}(p), F_{p}(p)),
\end{equation}
where $M_t$ is the reference line segment mask - essentially, loss is the average tangent angle discrepancy over the pixels that correspond to the ground truth line segments. 

\textbf{CNN Architecture}.
To predict the proposed feature map, we use a CNN of U-Net \cite{ronneberger2015u} family.
The architecture we use differs from the original one in the following simplifications.
We exploit padded convolutions providing the same input and output spatial sizes of convolutional layers, which allows not to crop the feature maps feeded to skip connections.
Instead of transposed convolutions we use bilinear upsampling.
We reduce the depth of encoder-decoder branches up to 3 maxpooling and 3 upsampling layers, correspondingly, and use fewer filters in convolutional blocks - 16, 32, 64, 128 filters per block (the number of blocks is greater than the number if maxpooling layers by one).
The resulting CNN has $\approx0.5$M trainable parameters and can run at 48 FPS on CPU and at 695 FPS on GPU (refer to section \ref{sec:setting} for benchmarking details).

\subsection{Line Segments Reconstruction}
Let us consider how the predicted segments mask $M$ and tangent field $F$ are converted into the desired output - a set of line segment $((x_1,y_1), (x_2,y_2)), ...$. This process has three steps:
firstly, the predicted features are coarsely segmented into lines and background (section ``Foreground segmentation'').
Secondly, the lines are finely segmented into several line support regions (LSRs) (section ``Region grouping'').
Finally, a line segment and its confidence is extracted from each LSR (section ``Line segments extraction'').

\textbf{Foreground segmentation}.
The first step of line segments reconstruction is to segment foreground (lines) from background (not lines).

We use a coarse-to-fine binarization approach by multiplying the masks of global thresholding $M(p) > \tau$ and local thresholding
\begin{equation}
     M(p) > \sum_{d\in K(p)}W_p(d)M(d)- \theta
\end{equation}
where $\theta$ - threshold, $K_{p}$ - a window centered at pixel $p$, $W_p(d)$ - Gaussian averaging weight.

\begin{figure}[htb]

\begin{minipage}[b]{.24\linewidth}
  \centering
  \centerline{\includegraphics[width=2.0cm]{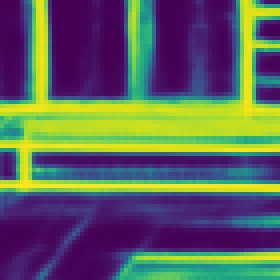}}
  \centerline{(a) Prediction}
\end{minipage}
\hfill
\begin{minipage}[b]{.24\linewidth}
  \centering
  \centerline{\includegraphics[width=2.0cm]{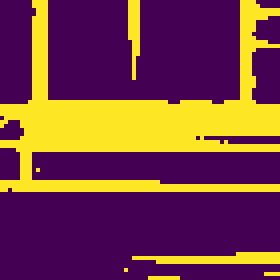}}
  \centerline{(b) Global}
\end{minipage}
\hfill
\begin{minipage}[b]{.24\linewidth}
  \centering
  \centerline{\includegraphics[width=2.0cm]{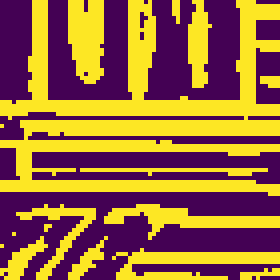}}
  \centerline{(c) Local}
\end{minipage}
\hfill
\begin{minipage}[b]{.24\linewidth}
  \centering
  \centerline{\includegraphics[width=2.0cm]{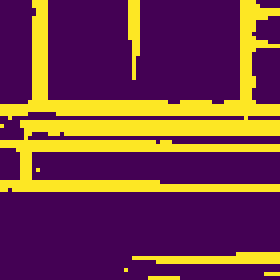}}
  \centerline{(d) Combined}
\end{minipage}
\caption{Binarization of CNN prediction. Global binarization joins close regions, while local binarization is noisy for the regions of near-zero intensity. The combined binarization is free of both drawbacks.}
\label{fig:binarization}
\end{figure}

Global thresholding with a small threshold gives a coarse extraction of line segments mask, but often incorrectly joins close - but separate - line segments.
On the contrary, local thresholding provides much finer local distinction of line segments, but can produce clumps of false positive detections in low intensity areas (Fig. \ref{fig:binarization}).
The combination of these binarizations by simple multiplication of the resulting masks allows to filter out false positives of both types and achieve better accuracy.

\textbf{Region grouping}.
The goal of this step (being a modification of a similar step of LSD algorithm \cite{von2008lsd}) is to split the foreground, binarized at the previous step, into narrow strip-like LSRs, one per true line segment.

Informally, we want neighbouring pixels $p_1, p_2$ to be assigned to one LSR, if the values of $M(p_1)$, $F(p_1)$ and $M(p_2)$, $F(p_2)$ are similar. The algorithm grows LSRs iteratively, starting from pixels with highest $M$ value and adding new pixels to the existing LSR which are geometrically close to the pixels and have similar features.

Let us formally introduce the similarity measure used to decide whether pixel $g$ is fit to be joined into a LSR. Let $R = p_1, p_2,...,p_n$ be the set of pixels of this LSR, $I_R = 1/n\cdot\sum_{p\in R} M(p)$ be the mean line segments' mask over it, and $\phi_R=\angle\sqrt{\sum_{p\in R} e^{2iF(p)}}$ - the average tangent field (here $\angle z = \atantwo(\cim(z) / \cre(z))$ is the phase of a complex number). Then the similarity function is given by
\begin{equation}
    d_{M,F}(g,R) = \rho^2(F(g),\phi_R)+\alpha(M(g)-I_R)^2
\end{equation}
The first term defines similarity of tangent field orientation (refer to Eq. (\ref{eq:angledist}) for details), the second - the similarity of line mask, $\alpha$ - weighting coefficient.
Given the distance function, LSRs are built with an iterative growing algorithm, presented in Algorithm \ref{alg:segmentation}.

\begin{algorithm}
\SetKwInOut{Input}{input}
\SetKwInOut{Output}{output}
\SetKwInOut{Parameter}{param}
\SetKwRepeat{Do}{do}{while}

\Input{Line heatmap $M\in [0,1]^{hw}$, \newline
 tangent field $F\in [0,\pi)^{hw}$}
\Output{Set of line support regions $\mathcal{R}=\{R\}$}
\Parameter{$\tau\in\mathbf{R}$ - distance threshold}
\Parameter{$n\in\mathbf{N}$ - minimum region size}
 $B\in \{0,1\}^{hw}$ - foreground segmentation of $M$\;
 $P=(p_i)$, such that $B(p)=1, M(p_i) \geq M(p_{i+1})$\;
 $P_{used} \leftarrow\varnothing$\;

 \For{$p\in P$}{
 \If{$p \in P_{used}$}{
  \textbf{continue}\;
    }
    $R \leftarrow p$\;
    
    \Do{$region\_updated$}{
        $region\_updated\leftarrow False$\;
        \For{$g\in N_8(R) \setminus P_{used}$}{
            \If{$d_{M,F}(g,R) < \tau$}{
                \textbf{Add} $R \leftarrow g$\;
                \textbf{Add} $P_{used} \leftarrow g$\;
                $region\_updated \leftarrow True$\;
            }
        }
    }
    \textbf{Add} $\mathcal{R}\leftarrow R$\;
 }
 \For{$R\in \mathcal{R}$}{
      \If{$|R| < n$}{
    remove $R$ from $\mathcal{R}$\;
    }
 }
 \caption{Region grouping algorithm. $d_{M,F}(p,R)$ - distance function between pixel $p$ and region $R$, $N_8(R)$ - 8-neighbours of region $R$.}
 \label{alg:segmentation}
\end{algorithm}

\textbf{Line segments extraction}. Each LSR $R = p_1, p_2,...,p_n$, should be converted into a line segment satisfying the following criteria. 
\begin{itemize}
    \item The segment goes through LSR's center of mass $p_\mu$
    \begin{equation}
    p_\mu=\frac{1}{\sum_{p\in R} M(p)} \sum_{p\in R} M(p)p.
    \end{equation}
    \item The segment is collinear with minor eigenvector $a$ of region's inertia tensor $I$ defines as follows
\begin{equation}
    I = \sum_{p\in R}M(p)I^*(p-p_\mu)
\end{equation}
\begin{equation}
    I^*(x,y)=\begin{pmatrix}
    y^2 & -xy\\
    -xy & x^2\\
    \end{pmatrix}
\end{equation}
    \item The segment spans the furthest LSR's points, projected onto axis $a$.
The segment's confidence is mean value of $M$ over the region $R$.
\end{itemize}
Line segment extraction is visualized in Figure \ref{fig:dist}b.

\section{Dataset}\label{sec:dataset}
In this section we analyze the issues of the existing line segment detection datasets and propose a dataset Wireframe-tiny++, a subset of Wireframe dataset \cite{huang2018learning} with refined annotations.

\subsection{The existing datasets}\label{sec:ds_exist}
\begin{figure*}[htbp]
\centering
\begin{minipage}[b]{0.3\linewidth}
  \centering
  \centerline{\includegraphics[width=\textwidth]{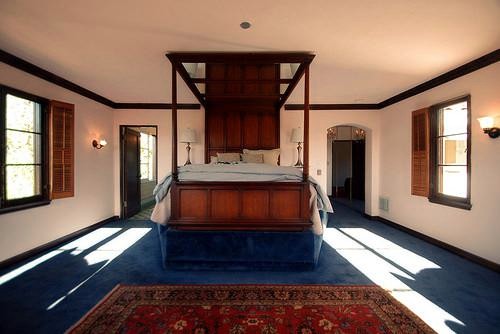}}
  \vspace{3pt}
  \centerline{\includegraphics[width=\textwidth]{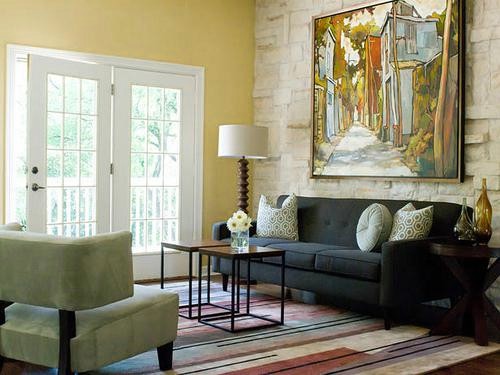}}
  \centerline{(a) Original image}
\end{minipage} 
\begin{minipage}[b]{.3\linewidth}
  \centering
  \centerline{\includegraphics[width=\textwidth]{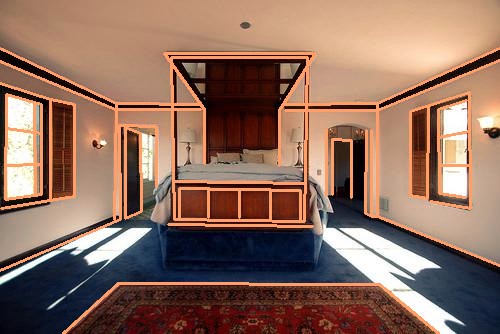}}
  \vspace{3pt}
  \centerline{\includegraphics[width=\textwidth]{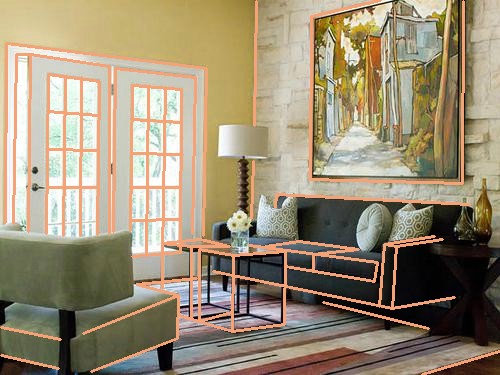}}
  \centerline{(b) Wireframe annotation}
\end{minipage}
\begin{minipage}[b]{.3\linewidth}
  \centering
  \centerline{\includegraphics[width=\textwidth]{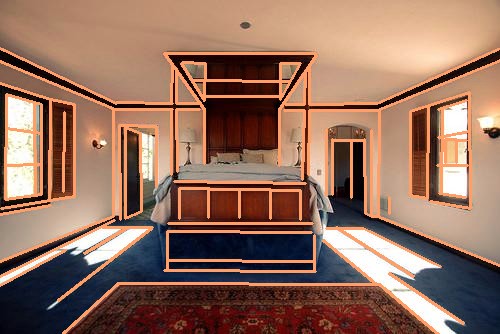}}
  \vspace{3pt}
  \centerline{\includegraphics[width=\textwidth]{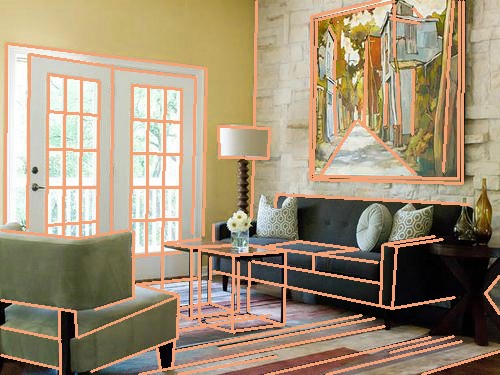}}
  \centerline{(c) Wireframe-tiny++ annotation}
\end{minipage}
\caption{An image from Wireframe dataset (a) and two versions of its annotation - original (b) and proposed (c).
Best viewed in color and zoom-in.}
\label{fig:ds_issue}
\end{figure*}

To the best of our knowledge, there are two widely-used public line segment detection datasets: Wireframe \cite{huang2018learning} and YorkUrban \cite{denis2008efficient}.
The former is composed of 5.000 train and 462 test images, the latter is composed of 120 test images.
The datasets contain both indoor and outdoor colour images of various man-made environments.
Some samples from the datasets are presented in Figures \ref{fig:ds_issue} and \ref{fig:vis_cmp}.
The datasets are annotated with a list of point pairs, representing line segments.

York dataset was annotated under so-called Manhattan world assumption \cite{coughlan2003manhattan}, which means that the annotated line segments are those aligned with the basis of some Cartesian coordinate system (specifically, with axes parallel to image sides), while the others are ignored. 
Wireframe dataset did not follow Manhattan world assumption and was annotated with line segments, from which ``meaningful geometric information of the scene can be extracted'' \cite{huang2018learning}, which also resulted in some salient line segments being not annotated. 

So, the ground truth labeling in these datasets is explicitly limited to some category of line segments - which means other categories of line segments are viewed as negatives. CNNs that are trained on these datasets (and/or with high accuracy on them) will have to systematically classify these categories of line segments as negatives and therefore - by design! - can't be viewed as general-purpose line detectors.

While such annotations can be useful to train and test some specific niche line segment detectors (e.g. for indoor robot navigation \cite{delage2007automatic}) we believe they are flawed as datasets for general purpose line segment detection.
Specifically, we would like to highlight the following problems (also illustrated in Fig. \ref{fig:ds_issue}b).

One problem is the inconsistency of the annotations - it is easily noticeable on strip-like objects  having two side line segments looking almost exactly alike - however one of them is annotated while the other is not.

Some categories of salient line segments are systematically not annotated - e.g. shadows and reflections. We believe the fact that these segments are not ``real'' physical objects should not be considered in the context of general purpose line segment detection and these segments should be annotated as well.


Finally, some line segments lying on the same straight line are falsely merged (vertical segments on the bed canopy's frame in Fig. \ref{fig:ds_issue}b).
It happens when a long line segment is intercepted by another object.
Although for some applications it could be desirable to avoid such a splitting and there are approaches to achieve that \cite{hamid2016lsm, yu2020plsd}, we believe, that for general-purpose detector splitting is the desired detector's behaviour.

\subsection{Wireframe-tiny++}\label{sec:wf_tiny}
To approach the covered issues with the existing datasets, we selected 20 random images from Wireframe test subset and reannotated them to make the annotations more accurate and consistent.
We call the selected subset of images with the original markup Wireframe-tiny, and the resulting dataset with enhanced annotations - Wireframe-tiny++.

Comparing to the original annotations, we mainly added unannotated segments, 9 per image on average.
Some segments are removed as undetectable.
Some segments are divided into several smaller segments due to occlusion.
The refined annotations are presented in Fig. \ref{fig:ds_issue}c.

\section{Experimental Setting}\label{sec:setting}

\textbf{Datasets}.
The proposed algorithm is trained and evaluated with the following datasets.
Wireframe dataset \cite{huang2018learning} consisting of 5000 training and 462 test images is used both to train and evaluate LSDNet.
Datasets YorkUrban \cite{denis2008efficient}, Wireframe-tiny and its reannotated version Wireframe-tiny++  (refer to the previous section for details), composed of 120 and 20 images correspondingly, are used solely for evaluation.

\textbf{Accuracy}.
To evaluate LSDNet accuracy we use standard \cite{xue2020holistically, zhou2019end} quality score $F^H = 100\cdot 2 \cdot p\cdot r / (p+r)$, where $p, r$ stand for precision and recall.
Multiplier $100$ is added for readability, making $F^H$ fall in $[0,100]$  range.
The score is evaluated pixel-wise by rasterizing both the predicted and the reference line segments.
A pixel of a predicted line segment is considered true positive, if its distance to a pixel of a reference segment does not exceed $1\%$ of image diagonal.
For evaluation we use $F^H$ implementation provided with L-CNN \cite{zhou2019end}.

Quality score $F^H$ was criticised \cite{zhou2019end} for being not sensitive towards overlapping and splitted line segments.
We consider such an insensibility is not critical: LSDNet can not produce overlapping segments by design, since line support regions (LSRs) can not overlap, and we did not observe a notable amount of splitted line segments for any CNN-based algorithm.

\textbf{Speed}.
CNN is benchmarked on Quadro GV100 GPU for comparison with other detectors.
Reconstruction algorithm is benchmarked on Core i5 9300hf CPU.
CNN and the reconstruction algorithm are benchmarked independently.
The speed of the latter depends on its input, we report the average speed over the dataset given the trained preprocessing network.

\textbf{Baselines}.
We  compare the proposed LSDNet with classical algorithm LSD \cite{von2008lsd} and several state-of-the-art CNN-based detectors L-CNN\cite{zhou2019end}, HAWP\cite{xue2020holistically}, TP-LSD\cite{huang2020tp}, M-LSD and M-LSD-tiny\cite{gu2021towards}.

The reported FPS for all CNN-based methods is cited as in \cite{gu2021towards}, where benchmarking was performed on Tesla V100 GPU with practically the same characteristics as the GPU used in our experiments.

The reported $F^H$ is also cited as in \cite{gu2021towards} for all methods except LSD and L-CNN \cite{zhou2019end}, for which it was reproduced by our means.
We tried to reproduce the stated quality measurements for other approaches with the help of their open-source implementations, but they appeared notably lower than the reported ones.
Therefore on datasets Wireframe-tiny and Wireframe-tiny++ we compare LSDNet only to L-CNN and LSD.

\textbf{Preprocessing}.
For LSDNet, all images are resized to $288\times288$, which appeared to be the optimal input shape in terms of speed-accuracy tradeoff.
Pixel intensities are simply converted from 8-bit unsigned integer to 32-bit floating-point with $1/255$ scaling coefficient.
During training, random horizontal and vertical flips and gamma correction are applied.

For baseline methods, the preprocessing from the corresponding paper is applied.
For LSD, we use $320\times320$ image shape.

\textbf{Hyperparameters}.
LSDNet is initialized with He uniform initialization\cite{he2015delving} and trained by Adam optimizer \cite{kingma2014adam} with $10^{-4}$ weight decay and $8$ images per batch for $180$ epochs.
The initial learning rate is $10^{-3}$ and is reduced by half if the value of loss function does not improve for $15$ epochs.

\textbf{Implementation}.
CNN training and inference is implemented in TensorFlow \cite{abadi2016tensorflow} and ONNX Runtime \cite{onnxruntime}, correspondingly.
Reconstruction algorithm is implemented in C++ with the help of OpenCV \cite{opencv_library}.

\section{Results and Analysis}

In this section we quantitatively and qualitatively analyze LSDNet performance and compare it to a wide range of state-of-the-art line segment detectors.
Please refer to Section \ref{sec:setting} for evaluation and comparison details.

\begin{figure}[htbp]
\centering
\includegraphics[width=0.49\textwidth]{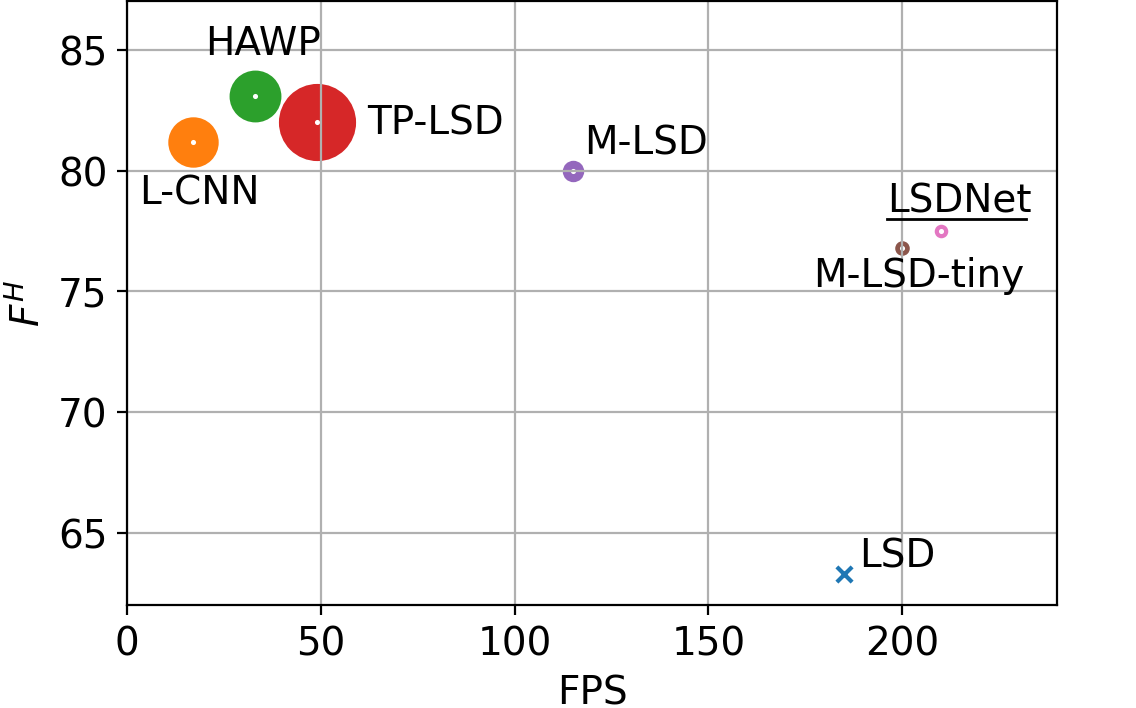}
\caption{Speed (FPS) and accuracy ($F^H$) (accuracy) comparisons of different line segment detectors. Size of circles indicates the number of trainable parameters.
LSD is marked as cross since it has no trainable parameters.
The proposed LSDNet outperforms all previous detectors in speed with $214$ FPS and outperforms the nearest fastest counterpart, M-LSD-tiny, both in speed and $F^H$ quality.}
\label{fig:vis_cmp}
\end{figure}

\begin{table}[htbp]
\centering
\begin{tabular}{| l |c c  | c c | c|}
 \hline
 \multirow{2}{*}{Method} & \multicolumn{2}{c|}{accuracy, $F^H$} & \multicolumn{3}{c|}{speed, FPS}\\

 & Wireframe & York & CNN & postproc & total\\
 \hline\hline
  LSD \cite{von2008lsd} & 63.3 & 59.0 & - & 185 & 185\\
  L-CNN \cite{zhou2019end} & 81.2 & 65.4  & 55 & 24 & 17 \\
  HAWP \cite{xue2020holistically} & \textbf{83.1} & \underline{66.3} & 55 & 82 & 33\\
  TP-LSD \cite{huang2020tp} & \underline{82.0} & \textbf{67.3} & 65 & 201 & 49\\
  M-LSD \cite{gu2021towards} & 80.0 & 64.2 & 132 & \underline{883} & 115\\
  M-LSD-tiny \cite{gu2021towards} & 76.8 & 61.9 & \underline{241} &\textbf{1203} &\underline{200} \\
  LSDNet (proposed) &  77.5 & 64.6 & \textbf{695} & 301 & \textbf{214} \\
  \hline
\end{tabular}
\caption{Detection accuracy on datasets Wireframe and York Urban. Bold and underlined values stand for top-1 and top-2, correspondingly.
Column ``postproc'' shows the speed of postprocessing the CNN's prediction.}
\label{tab:accuracy_wf_york}
\end{table}

\begin{table}[htbp]
\centering
\begin{tabular}{| l | c c c |}
 \hline
 \multirow{2}{*}{Method} & \multicolumn{3}{c|}{detection accuracy, $F^H$} \\
 & Wireframe & Wireframe-tiny & Wireframe-tiny++ \\
 \hline\hline
  LSD\cite{von2008lsd} & 63.3 &  69.0 & 73.2 \\
  L-CNN \cite{zhou2019end} & \textbf{81.2} & \textbf{83.6} & \textbf{85.5} \\
  LSDNet (proposed) &  \underline{77.5} &  \underline{82.0} & \underline{85.3} \\
  \hline
\end{tabular}
\caption{Detection accuracy on datasets Wireframe, Wireframe-tiny, Wireframe-tiny++ datasets. Bold and underlined values stand for top-1 and top-2, correspondingly.}
\label{tab:accuracy_wf_tiny}
\end{table}

\begin{figure*}[h!]
\centering
\centerline{\includegraphics[width=\textwidth]{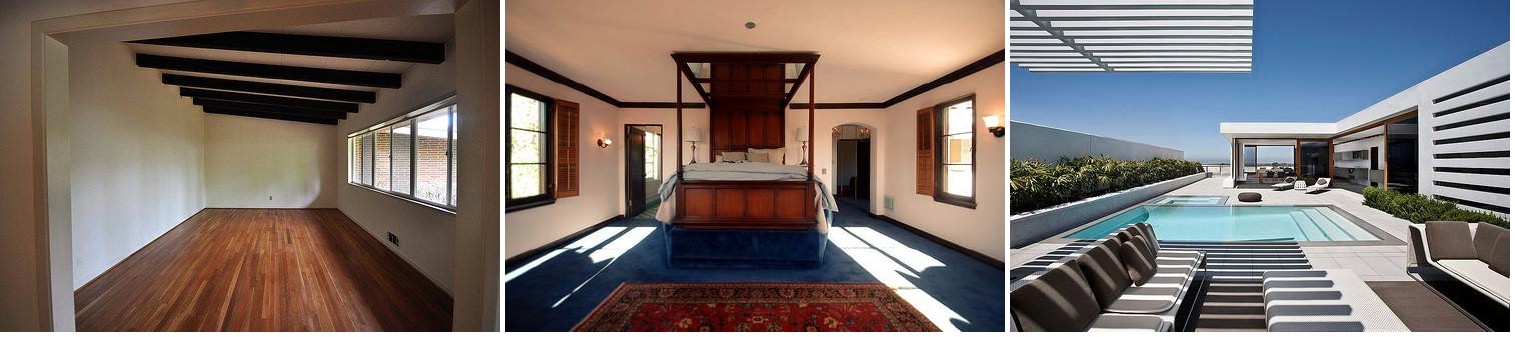}}
\centerline{\includegraphics[width=\textwidth]{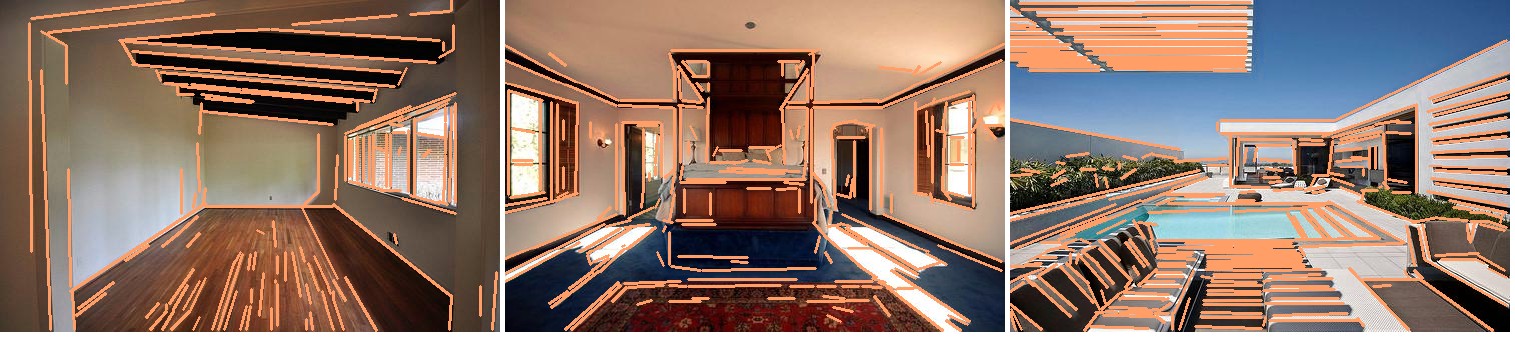}}
\centerline{\includegraphics[width=\textwidth]{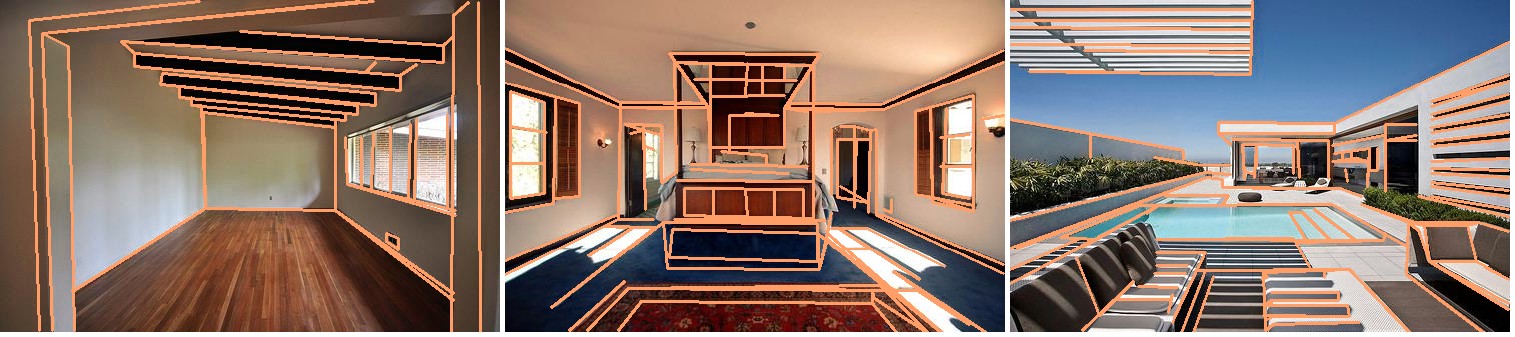}}
\centerline{\includegraphics[width=\textwidth]{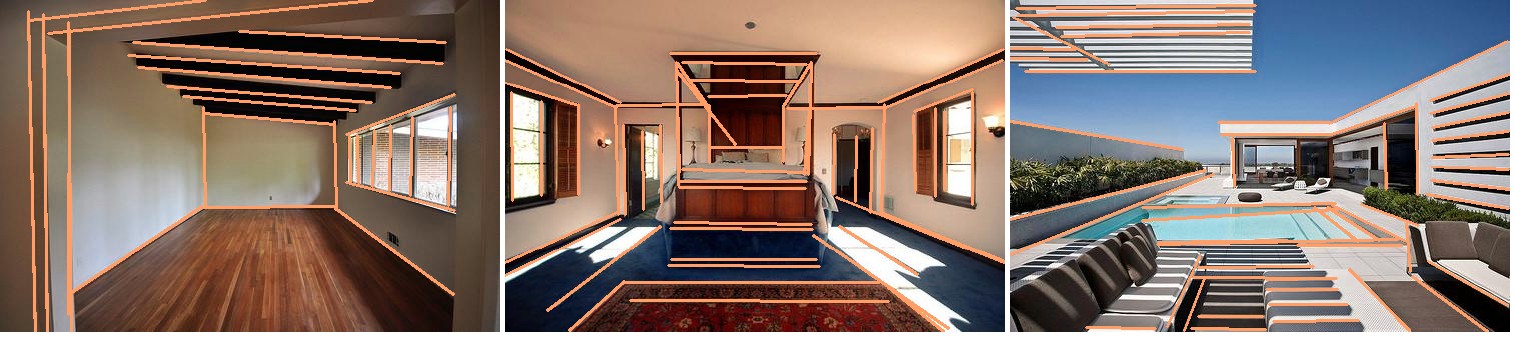}}
\centerline{\includegraphics[width=\textwidth]{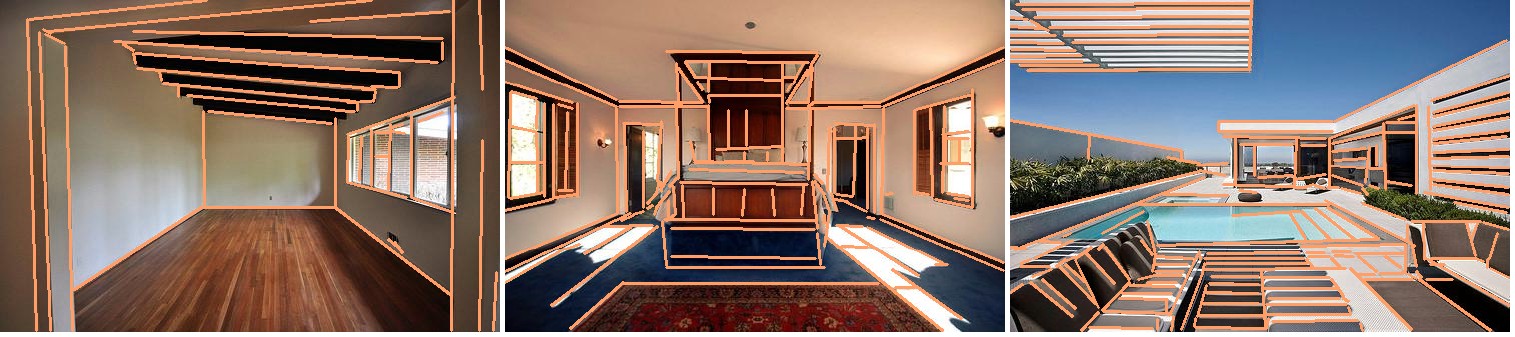}}
\caption{Qualitative evaluation.
From left to right - images from Wireframe dataset of varying number of salient line segments.
From top to bottom - original image, detection results from LSD \cite{von2008lsd}, the model-based detector; HAWP \cite{xue2020holistically}, the most accurate CNN-based detector; M-LSD-tiny \cite{gu2021towards}, the fastest CNN-based detector except for the proposed one; the proposed LSDNet.
Best viewed in color and zoom-in.}
\label{fig:example}
\end{figure*}

\textbf{Public datasets}.
Table \ref{tab:accuracy_wf_york} and Figure \ref{fig:vis_cmp} summarize the results on Wireframe and York Urban datasets.
It shows that LSDNet achieves state-of-the-art inference speed of $695$ FPS for standalone CNN and $214$ FPS for overall detector alongside competitive accuracy with $77.5$ $F^H$ and $64.6$ $F^H$ on Wireframe and York Urban datasets, correspondingly.

In comparison with the nearest fastest counterpart, M-LSD-tiny, LSDNet outperforms it both in quality and speed with the absolute increase in accuracy of $+1.0$ $F^H$ on Wireframe, $+2.7$ $F^H$ on York Urban and $+14$ FPS speed-up.

Compared to the fastest detector outperformimg LSDNet in accuracy on Wireframe dataset, M-LSD, the proposed approach is approximately two times faster with $214$ FPS against $115$ FPS.
The fastest algorithm to outperform  LSDNet on both datasets is TP-LSD, which is approximately four times slower with $49$ FPS.

\textbf{Custom datasets}.
Table \ref{tab:accuracy_wf_tiny} summarizes the results on Wireframe dataset, its subset Wireframe-tiny and its reannotated version Wireframe-tiny++.
Please refer to section \ref{sec:wf_tiny} for details.

On Wireframe-tiny++, all the detectors demonstrate higher accuracy than those on Wireframe-tiny.
Since these datasets are composed of the same images and differ only in annotations, such a consistent accuracy growth indicates that Wireframe-tiny++ annotation is more suitable for the problem of general purpose line segment detection.

All the approaches, being arranged by $F^H$, show the same relative order on all the datasets, but the absolute differences change significantly.
The gap between LSDNet and L-CNN has shrinked from $3.7$ $F^H$ on Wireframe to negligible $0.2$ $F^H$ on Wireframe-tiny++.
We believe it could be explained by different learning capacity of detectors' CNNs.
Expressive L-CNN with 9.8M parameters managed to learn the subtle notion of a line segment implied by Wireframe train dataset annotation (discussed in Sec. \ref{sec:ds_exist}); whereas lightweight LSDNet with only 0.5M parameters learned the general line segment detection with no capacity to learn the subtle details.
It made L-CNN good for wireframe-like detection problems with the goal to detect line segments, from which ``meaningful geometric information of the scene can be extracted''.
But it could possess confusing properties in terms of general-purpose line segment detection, making LSDNet a better choice in such a case.

\textbf{Qualitative results}.
Qualitative comparison of LSDNet to other line segment detectors is illustrated in Figure \ref{fig:example}.
In this section we refer to LSD and LSDNet as LSR-based and to HAWP and M-LSD-tiny as endpoint-based detectors, since the methods within these groups demonstrate similar behaviour.

Endpoint-based detectors demonstrate the selectivity of line segments, which could not be attributed to overall segments' saliency.
This effect is mostly notable in the foreground in the right column in Figure \ref{fig:example}.
LSR-based LSD detects all the shadows and the carpet in front of the sofa, while it can't ``see'' floor tiles due to their low contrast.
$\text{LSDNet}$ detects all the shadows, the carpet and the floor tiles.
Whereas endpoint-based detectors HAWP and M-LSD-tiny detect these objects poorly, but at the same time they detect way less salient segments in the background.
We believe such a selectivity could be attributed to the combination of high expressive power of the underlying CNN and annotation inconsistencies of train dataset, discussed in Section \ref{sec:ds_exist}.
This effect could be undesirable in an application requiring that very class of segments, which is missed by endpoint-based detectors.

Another interesting difference between the detectors' groups occurs due to the types of misdetections.
In terms of a quality measure, LSR-based LSD and LSDNet can detect false positives, typically corresponding to line segment-like patterns on highly structured image regions, and miss some annotated segments (false negatives), which are usually poorly visible.
These errors could be, at least partially, attributed to the ill-posed nature of the task.
The endpoints-based methods are also prone to miss poorly visible segments, and possess an advantage of not detecting false positives on structured image regions.
However, an potential drawback of endpoints-based detectors is that they can produce hard false positives --- line segments of high confidence score with no evidence of a true line segment in an image.
We believe the reason for hard false positives is a classification error of line segment verification module.
An example could be seen in the middle column in Figure \ref{fig:example}, please note the salient diagonal segments in the bedhead (fourth row) and right part of the carpet (third row).
This issue is to be approached prior to successful exploitation of an endpoint-based detector.

\section{Conclusion}
In this study we introduce a fast and accurate line segment detector $\text{LSDNet}$.
The detector is composed of a lightweight encoder-decoder CNN, which predicts line segment heatmap and tangent field, and a postprocessing module -- a modification of the famous LSD algorithm.
When benchmarked on the traditional Wireframe dataset against several SOTA methods, LSDNet shows the highest FPS of $214$ -- though it achieves detection accuracy of $78$ $F^H$  -- lower than the best methods (82 and 83.1 $F^H$).

However, we speculate that this gap in detection accuracy is primarily caused by the imperfections of the dataset rather than the network itself. We analyze the commonly used line segment detection datasets -- Wireframe and York Urban -- and point out numerous and significant inconsistencies in their annotation. By carefully reannotating a part of the Wireframe test dataset, we show that (i) all detectors demonstrate better quality on improved annotations (without any re-training), which indicates that the refined annotations correlate better with the notion of correct line segment detection, (ii) the gap between accuracies of our detector and others is reduced to almost non-existent - with our method being the fastest.

\section{Acknowledgements}
We would like to thank Marina Tepliakova for making the illustrations; Alexey Savchik and Veniamin Blinov for reviewing the early versions of the manuscript; Dmitry Nikolaev for his strong belief in the power of fusing model-based algorithms with light-weight neural networks, which inspired this work.

\balance
\bibliographystyle{IEEEtran}
\bibliography{main}

\begin{thebibliography}{10}
\providecommand{\url}[1]{#1}
\csname url@samestyle\endcsname
\providecommand{\newblock}{\relax}
\providecommand{\bibinfo}[2]{#2}
\providecommand{\BIBentrySTDinterwordspacing}{\spaceskip=0pt\relax}
\providecommand{\BIBentryALTinterwordstretchfactor}{4}
\providecommand{\BIBentryALTinterwordspacing}{\spaceskip=\fontdimen2\font plus
\BIBentryALTinterwordstretchfactor\fontdimen3\font minus
  \fontdimen4\font\relax}
\providecommand{\BIBforeignlanguage}[2]{{%
\expandafter\ifx\csname l@#1\endcsname\relax
\typeout{** WARNING: IEEEtran.bst: No hyphenation pattern has been}%
\typeout{** loaded for the language `#1'. Using the pattern for}%
\typeout{** the default language instead.}%
\else
\language=\csname l@#1\endcsname
\fi
#2}}
\providecommand{\BIBdecl}{\relax}
\BIBdecl

\bibitem{andrade2018novel}
D.~C. Andrade, F.~Bueno, F.~R. Franco, R.~A. Silva, J.~H.~Z. Neme, E.~Margraf,
  W.~T. Omoto, F.~A. Farinelli, A.~M. Tusset, S.~Okida \emph{et~al.}, ``A novel
  strategy for road lane detection and tracking based on a vehicle’s forward
  monocular camera,'' \emph{IEEE Transactions on Intelligent Transportation
  Systems}, vol.~20, no.~4, pp. 1497--1507, 2018.

\bibitem{hoang2016road}
T.~M. Hoang, H.~G. Hong, H.~Vokhidov, and K.~R. Park, ``Road lane detection by
  discriminating dashed and solid road lanes using a visible light camera
  sensor,'' \emph{Sensors}, vol.~16, no.~8, p. 1313, 2016.

\bibitem{shipitko2020linear}
O.~Shipitko, V.~Kibalov, and M.~Abramov, ``Linear features observation model
  for autonomous vehicle localization,'' 2020.

\bibitem{ceron2014power}
A.~Ceron, F.~Prieto \emph{et~al.}, ``Power line detection using a circle based
  search with uav images,'' in \emph{2014 International Conference on Unmanned
  Aircraft Systems (ICUAS)}.\hskip 1em plus 0.5em minus 0.4em\relax IEEE, 2014,
  pp. 632--639.

\bibitem{cao2012automatic}
Y.~Cao and L.~Yan, ``Automatic road network extraction from uav image in
  mountain area,'' in \emph{2012 5th International Congress on Image and Signal
  Processing}.\hskip 1em plus 0.5em minus 0.4em\relax IEEE, 2012, pp.
  1024--1028.

\bibitem{shemiakina2018method}
J.~Shemiakina, A.~Zhukovsky, and D.~Nikolaev, ``The method for homography
  estimation between two planes based on lines and points,'' in \emph{Tenth
  International Conference on Machine Vision (ICMV 2017)}, vol. 10696.\hskip
  1em plus 0.5em minus 0.4em\relax International Society for Optics and
  Photonics, 2018, p. 106961G.

\bibitem{zhukovsky2017segments}
A.~Zhukovsky, D.~Nikolaev, V.~Arlazarov, V.~Postnikov, D.~Polevoy,
  N.~Skoryukina, T.~Chernov, J.~Shemiakina, A.~Mukovozov, I.~Konovalenko
  \emph{et~al.}, ``Segments graph-based approach for document capture in a
  smartphone video stream,'' in \emph{2017 14th IAPR International Conference
  on Document Analysis and Recognition (ICDAR)}, vol.~1.\hskip 1em plus 0.5em
  minus 0.4em\relax IEEE, 2017, pp. 337--342.

\bibitem{von2008lsd}
R.~G. Von~Gioi, J.~Jakubowicz, J.-M. Morel, and G.~Randall, ``Lsd: A fast line
  segment detector with a false detection control,'' \emph{IEEE transactions on
  pattern analysis and machine intelligence}, vol.~32, no.~4, pp. 722--732,
  2008.

\bibitem{akinlar2011edlines}
C.~Akinlar and C.~Topal, ``Edlines: Real-time line segment detection by edge
  drawing (ed),'' in \emph{2011 18th IEEE International Conference on Image
  Processing}.\hskip 1em plus 0.5em minus 0.4em\relax IEEE, 2011, pp.
  2837--2840.

\bibitem{guerreiro2012connectivity}
R.~F. Guerreiro and P.~M. Aguiar, ``Connectivity-enforcing hough transform for
  the robust extraction of line segments,'' \emph{IEEE Transactions on Image
  Processing}, vol.~21, no.~12, pp. 4819--4829, 2012.

\bibitem{xu2014accurate}
Z.~Xu, B.-S. Shin, and R.~Klette, ``Accurate and robust line segment extraction
  using minimum entropy with hough transform,'' \emph{IEEE Transactions on
  Image Processing}, vol.~24, no.~3, pp. 813--822, 2014.

\bibitem{hamid2016lsm}
N.~Hamid and N.~Khan, ``Lsm: perceptually accurate line segment merging,''
  \emph{Journal of Electronic Imaging}, vol.~25, no.~6, p. 061620, 2016.

\bibitem{yu2020plsd}
Q.~Yu, G.~Xu, Y.~Cheng, and Z.~H. Zhu, ``Plsd: A perceptually accurate line
  segment detection approach,'' \emph{IEEE Access}, vol.~8, pp.
  42\,595--42\,607, 2020.

\bibitem{xue2019learning}
N.~Xue, S.~Bai, F.~Wang, G.-S. Xia, T.~Wu, and L.~Zhang, ``Learning attraction
  field representation for robust line segment detection,'' in
  \emph{Proceedings of the IEEE Conference on Computer Vision and Pattern
  Recognition}, 2019, pp. 1595--1603.

\bibitem{xue2020holistically}
N.~Xue, T.~Wu, S.~Bai, F.~Wang, G.-S. Xia, L.~Zhang, and P.~H. Torr,
  ``Holistically-attracted wireframe parsing,'' in \emph{Proceedings of the
  IEEE/CVF Conference on Computer Vision and Pattern Recognition}, 2020, pp.
  2788--2797.

\bibitem{lin2020deep}
Y.~Lin, S.~L. Pintea, and J.~C. van Gemert, ``Deep hough-transform line
  priors,'' \emph{arXiv preprint arXiv:2007.09493}, 2020.

\bibitem{huang2020tp}
S.~Huang, F.~Qin, P.~Xiong, N.~Ding, Y.~He, and X.~Liu, ``Tp-lsd: Tri-points
  based line segment detector,'' \emph{arXiv preprint arXiv:2009.05505}, 2020.

\bibitem{zhou2019end}
Y.~Zhou, H.~Qi, and Y.~Ma, ``End-to-end wireframe parsing,'' in
  \emph{Proceedings of the IEEE International Conference on Computer Vision},
  2019, pp. 962--971.

\bibitem{huang2019wireframe}
K.~Huang and S.~Gao, ``Wireframe parsing with guidance of distance map,''
  \emph{IEEE Access}, vol.~7, pp. 141\,036--141\,044, 2019.

\bibitem{gu2021towards}
G.~Gu, B.~Ko, S.~Go, S.-H. Lee, J.~Lee, and M.~Shin, ``Towards real-time and
  light-weight line segment detection,'' \emph{arXiv preprint
  arXiv:2106.00186}, 2021.

\bibitem{bochkovskiy2020yolov4}
A.~Bochkovskiy, C.-Y. Wang, and H.-Y.~M. Liao, ``Yolov4: Optimal speed and
  accuracy of object detection,'' \emph{arXiv preprint arXiv:2004.10934}, 2020.

\bibitem{tian2019fcos}
Z.~Tian, C.~Shen, H.~Chen, and T.~He, ``Fcos: Fully convolutional one-stage
  object detection,'' in \emph{Proceedings of the IEEE/CVF international
  conference on computer vision}, 2019, pp. 9627--9636.

\bibitem{huang2018learning}
K.~Huang, Y.~Wang, Z.~Zhou, T.~Ding, S.~Gao, and Y.~Ma, ``Learning to parse
  wireframes in images of man-made environments,'' in \emph{Proceedings of the
  IEEE Conference on Computer Vision and Pattern Recognition}, 2018, pp.
  626--635.

\bibitem{denis2008efficient}
P.~Denis, J.~H. Elder, and F.~J. Estrada, ``Efficient edge-based methods for
  estimating manhattan frames in urban imagery,'' in \emph{European conference
  on computer vision}.\hskip 1em plus 0.5em minus 0.4em\relax Springer, Berlin,
  Heidelberg, 2008, pp. 197--210.

\bibitem{burns1986extracting}
J.~B. Burns, A.~R. Hanson, and E.~M. Riseman, ``Extracting straight lines,''
  \emph{IEEE transactions on pattern analysis and machine intelligence}, no.~4,
  pp. 425--455, 1986.

\bibitem{teplyakov2020line}
L.~Teplyakov, K.~Kaymakov, E.~Shvets, and D.~Nikolaev, ``Line detection via a
  lightweight cnn with a hough layer,'' \emph{arXiv preprint arXiv:2008.08884},
  2020.

\bibitem{li2021lidar}
Z.~Li, F.~Wang, and N.~Wang, ``Lidar r-cnn: An efficient and universal 3d
  object detector,'' \emph{arXiv preprint arXiv:2103.15297}, 2021.

\bibitem{ronneberger2015u}
O.~Ronneberger, P.~Fischer, and T.~Brox, ``U-net: Convolutional networks for
  biomedical image segmentation,'' in \emph{International Conference on Medical
  image computing and computer-assisted intervention}.\hskip 1em plus 0.5em
  minus 0.4em\relax Springer, Cham, 2015, pp. 234--241.

\bibitem{coughlan2003manhattan}
J.~M. Coughlan and A.~L. Yuille, ``Manhattan world: Orientation and outlier
  detection by bayesian inference,'' \emph{Neural computation}, vol.~15, no.~5,
  pp. 1063--1088, 2003.

\bibitem{delage2007automatic}
E.~Delage, H.~Lee, and A.~Y. Ng, ``Automatic single-image 3d reconstructions of
  indoor manhattan world scenes,'' in \emph{Robotics Research}.\hskip 1em plus
  0.5em minus 0.4em\relax Springer, Berlin, Heidelberg, 2007, pp. 305--321.

\bibitem{he2015delving}
K.~He, X.~Zhang, S.~Ren, and J.~Sun, ``Delving deep into rectifiers: Surpassing
  human-level performance on imagenet classification,'' in \emph{Proceedings of
  the IEEE international conference on computer vision}, 2015, pp. 1026--1034.

\bibitem{kingma2014adam}
D.~P. Kingma and J.~Ba, ``Adam: A method for stochastic optimization,''
  \emph{arXiv preprint arXiv:1412.6980}, 2014.

\bibitem{abadi2016tensorflow}
M.~Abadi, P.~Barham, J.~Chen, Z.~Chen, A.~Davis, J.~Dean, M.~Devin,
  S.~Ghemawat, G.~Irving, M.~Isard \emph{et~al.}, ``Tensorflow: A system for
  large-scale machine learning,'' in \emph{12th symposium on operating systems
  design and implementation}, 2016, pp. 265--283.

\bibitem{onnxruntime}
\BIBentryALTinterwordspacing
``Optimize and accelerate machine learning inferencing and training.''
  [Online]. Available: \url{https://www.onnxruntime.ai/}
\BIBentrySTDinterwordspacing

\bibitem{opencv_library}
\BIBentryALTinterwordspacing
G.~Bradski, ``Opencv library,'' Accessed on: Apr. 22, 2022. [Online].
  Available: \url{https://opencv.org/}
\BIBentrySTDinterwordspacing

\end{thebibliography}

\newpage
\begin{IEEEbiography}
[{\includegraphics[width=1in,height=1.25in,clip,keepaspectratio]{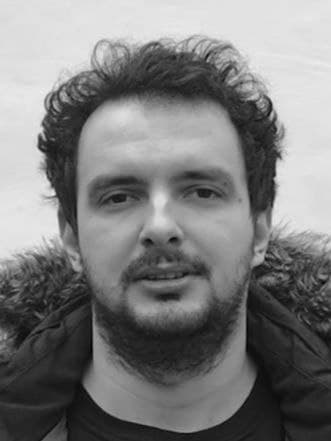}}]{Lev Teplyakov}
received the B.S. and M.S. degrees at Moscow Institute for Physics and Technology, Moscow, Russia in 2017 and 2019, correspondingly.
We is currently pursuing the Ph.D. degree at Institute for Information Transmission Problems, Moscow, Russia.

His master`s thesis considered the problem of training cascade CNN classifiers, which process an image with fast or slow branches, depending on the complexity of an image.
His research interests include video analytics, action recognition, deep learning for object detection, fast neural networks.
\end{IEEEbiography}

\vfill
\begin{IEEEbiography}
    [{\includegraphics[width=1in,height=1.25in,clip,keepaspectratio]{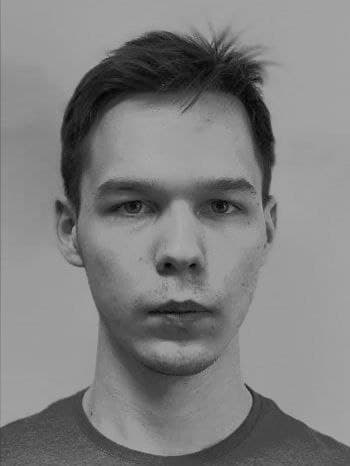}}]{Leonyd Erlygin}
graduated from Moscow Institute for Physics and Technology and received his bachelor degree in 2021.
Now he is a Master`s student at Institute for Information Transmission Problems, Moscow, Russia.

His research interests include CNN-based object detection and classification, with the focus on generation of synthetic training datasets with the help of data augmentation and 3d modelling.
\end{IEEEbiography}

\begin{IEEEbiography}
    [{\includegraphics[width=1in,height=1.25in,clip,keepaspectratio]{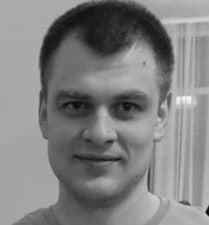}}]{Evgeny Shvets}
graduated from Moscow Institute for Physics and Technology, Moscow, Russia and received his Ph.D. degree in Technology in 2017 in Institute for Information Transmission Problems, Moscow, Russia.

His Ph.D. thesis focused on distributed control of a swarm for area surveillance.
His research interests include image processing, image registration and deep learning, including generation of synthetic datasets for neural network training.
\end{IEEEbiography}

\EOD

\end{document}